\documentclass[paper=letter]{scrartcl}
\usepackage[latin1]{inputenc}		\usepackage[T1]{fontenc} 

\date{}

\title{Warm Starting Bayesian Optimization\thanks{The authors were partially supported by NSF CAREER CMMI-1254298, NSF CMMI-1536895, NSF IIS-1247696, AFOSR FA9550-12-1-0200, AFOSR FA9550-15-1-0038, and AFOSR FA9550-16-1-0046.}}

\usepackage{authblk}
\author{Matthias Poloczek}
\author{Jialei Wang} 
\author{Peter I.\ Frazier}
\affil{School of Operations Research and Information Engineering\\Cornell University\\
	\texttt{\{poloczek,jw865,pf98\}@cornell.edu}}

\usepackage[cmex10]{amsmath}
\usepackage{amssymb,amsthm,url}
\usepackage{bbm,bm}					

\usepackage{microtype}

\usepackage{xcolor}					\usepackage{hyperref}				\definecolor{pololinkfarbe}{rgb}{0,0,0}
\hypersetup{
colorlinks,citecolor=pololinkfarbe,filecolor=pololinkfarbe,linkcolor=pololinkfarbe,urlcolor=pololinkfarbe}

\usepackage{tikz} 					\usetikzlibrary{arrows}				\usetikzlibrary{positioning}
\definecolor{darkgreen}{rgb}{0.0,0.6,0.2}

\usepackage{graphicx}
\usepackage{complexity}

\usepackage{xspace}
\renewcommand{\E}{\ensuremath{\mathbb{E}}}		
\newcommand{\argmax}{\operatornamewithlimits{argmax}}

\newcommand{\I}{\ensuremath{\mathbbm{1}}}
\newcommand{\domain}{\ensuremath{{\mathbb{D}}}\xspace}
\newcommand{\discretedomain}{\ensuremath{{\mathbb{A}}}\xspace}
\newcommand{\EGO}{\texttt{EGO}\xspace}
\newcommand{\Task}{\ensuremath{{\mathrm{T}}}\xspace}

\newcommand{\EI}{\texttt{EI}\xspace}

\newcommand{\KG}{\texttt{KG}\xspace}

\newcommand{\WSKG}{\texttt{wsKG}\xspace}
\newcommand{\ATO}{\texttt{ATO}\xspace}

\allowdisplaybreaks				
\overfullrule=5pt				\definecolor{pololinkfarbe}{rgb}{0,0,0.8}

\begin{document}							
												
\maketitle

\section*{ABSTRACT}

We develop a framework for warm-starting Bayesian optimization, that reduces the solution time required to solve an optimization problem that is one in a sequence of related problems.  This is useful when optimizing the output of a stochastic simulator that fails to provide derivative information, for which Bayesian optimization methods are well-suited.  Solving sequences of related optimization problems arises when making several business decisions using one optimization model and input data collected over different time periods or markets.  While many gradient-based methods can be warm started by initiating optimization at the solution to the previous problem, this warm start approach does not apply to Bayesian optimization methods, which carry a full metamodel of the objective function from iteration to iteration.  Our approach builds a joint statistical model of the entire collection of related objective functions, and uses a value of information calculation to recommend points to evaluate.

\newpage

\section{INTRODUCTION}
\label{section_introduction}
Businesses that use optimization to make operational decisions often solve collections of related optimization problems.
Consider three examples:

\begin{itemize}
\item First, consider a ride-sharing company, such as Uber or Lyft, that makes real-time decisions about how to assign requests for rides to drivers that may fulfill those requests. 
Such a company might have a single algorithm for making these decisions: its performance depends crucially on tunable parameters.
Given a particular distributional forecast of demand patterns, tuned to the particular next time period, and a simulator to estimate the performance of the algorithm using this forecast, this company might wish to optimize the parameters every few hours to find a configuration that would best achieve their goals in that time interval.
Such distributional forecasts would be likely to change across hours within a day and across days within a week due to seasonality, and would also vary from week to week as distributional forecasts include more recent historical data due to changing weather, sporting events, etc.
 
\item Second, consider a company managing inventory: to avoid overstocking or outages, an inventory management system has to decide what quantity of each item to hold in stock and how to set the threshold levels to order supplies. To maximize the profit, not only the current market prices have to be taken into account but also holding costs, delivery times affecting the lead times, and patterns in the demand.
Thus, the company would re-optimize the parameters of its inventory management system on a regular schedule and whenever any of the key factors change.

\item Third, consider a business that relies on machine learning models to predict the behavior of their clients. Suppose that such a model is used to predict ``click-rates'' for ads displayed on a collection of webpages. These predictions are subsequently used to decide which ad is to be displayed to each visitor, based on the user's profile but also on the targets agreed on in the advertising campaigns.
To obtain the most accurate predictions, the company would want to retrain the parameters of their model frequently based on the latest historical data, some of which typically have to be fit by a derivative-free method.
\end{itemize}

The common characteristic of all these problems is that we are to solve very similar problems over and over again.
To squeeze the most out of the real time availability of data, we would wish to ``shortcut'' the optimization process, thereby saving time and money and most importantly gaining an edge over competitors, \emph{without} negatively affecting the quality of solutions.
One way to avoid a ``cold start'', i.e.\ starting from scratch each time, is to exploit knowledge obtained while previously tackling related problems. 
This comprises not only the final solutions that were actually used in the application, but also all those evaluated in the respective optimization processes.

The idea of using previous solutions to speed up the optimization process is ubiquitous for applications in mathematical programming: in particular, the number of iterations that interior points methods for linear programming require to obtain an almost optimal solution can be reduced by a warm start, e.g., see~\cite{yw02,gw04,jw08}. 
Typically this is achieved simply by using the solution to a previous optimization problem as the starting point when solving the current problem.  This same technique can be used when using stochastic gradient ascent to solve optimization via simulation problems whose objective function evaluations provide stochastic gradients.  However, when evaluations of the objective function are derivative-free, as they commonly are when optimizing complex simulators, many of the optimization methods that are most effective for solving a single optimization problem instance lack the structure that would permit warm starting them using this technique.

In this article, we create the ability to warm start Bayesian optimization methods.  Bayesian optimization methods are effective for solving optimization via simulation problems in which the objective does not provide gradients, is expensive to evaluate, and takes a small number (< 20) of real- or integer-valued inputs.  
For an overview, see Bayesian optimization tutorials and surveys \cite{fw16,BrCoFr10,kleijnen2014simulation}.
Bayesian optimization methods belong to a wider set of optimization via simulation methods that leverage Bayesian statistics and value of information calculations.  See, e.g., the surveys of \cite{chick2000,Frazier2011b}.

Bayesian optimization methods iteratively construct a Gaussian process metamodel over the objective function and use this metamodel to decide whether to sample next.  Because the information carried from iteration to iteration is not just a single point but the full metamodel, simply carrying the solution from a previous problem as a starting point does not provide a warm starting capability to Bayesian optimization, necessitating a more sophisticated approach.  While one can partially warm start Bayesian optimization by performing function evaluations of the current problem at solutions to previous problems, this fails to take full advantage of the set of previous evaluations at non-optimal points.

In this article, we propose a Bayesian framework that captures the observed relationship of the current black-box function to previous ones. This enables us to perform a ``warm start'' and hence to zoom in on an optimal solution faster.
Our approach puts an emphasis on conceptual simplicity and wide applicability. In particular, we will model the tasks and their relationships by a Gaussian Process, whose covariance structure is tailored to capture the differences of the current task and the previous ones.

To optimize the current objective, we iteratively sample the search space, using the Knowledge Gradient acquisition criterion proposed by Frazier, Powell, and Dayanik~\cite{FrPoDa_Correlated}; besides its one-step Bayes optimality, this criterion has the advantage that it has a simple form and can be computed efficiently. However, our statistical model is independent of the acquisition function and can be combined with others if that seems beneficial for the application under consideration.

\medskip
\noindent{\textbf{Related Literature.}}
The previous work most directly relevant to warm starting Bayesian optimization and optimization of expensive functions more generally comes from the machine learning literature, where hyper-parameters in machine learning models are fit by solving optimization problems with expensive derivative-free evaluations.

Within this literature, the most relevant previous work is due to Swersky, Snoek, and Adams~\cite{ssa13}, who developed a warm start algorithm to tune hyper-parameters of machine learning models using Bayesian Optimization. 
They demonstrate that their algorithm can speed up the fine-tuning process for hyper-parameters of convolutional neural networks and logistic regression for image classification.
Their approach is similar to ours, as they also construct a Gaussian Process model to quantify the relationships between all previous instances and then use an acquisition function, in their case 
Entropy Search,
to optimize the hyper-parameters for the new instance.

We believe that our covariance kernel has the following advantages.
Their model assumes the covariance of each pair of tasks to be uniform over the whole domain. Moreover, they require that all tasks are positively correlated. 
Our covariance function does not impose these restrictions and in particular allows for the relationship of tasks to vary over the domain. Additionally, we quantify biases between the objective functions of different tasks. 
Finally, the inference of the parameters of their kernel requires an expensive sampling-based step, involving slice sampling, whereas our approach performs inference in closed form.

In other related work from machine learning, Feurer, Springenberg, and Hutter~\cite{fsh14} proposed a meta-learning procedure that improves the performance of a machine learning algorithm by selecting a small set of start-up configurations. The selection procedure explores the new task and then ranks previous tasks (and their corresponding optimal solutions) based on a metric that relies on an evaluation of the  ``metafeatures''  that grasp the characteristics of each dataset. The authors demonstrate that their approach can speed up the process of optimizing hyper-parameters for Support Vector Machines and Algorithm Selection models.  This approach differs from ours in three ways:  first, the metafeatures and distance functions it develops are specific to machine learning, and extending this approach in a generic way to optimization via simulation problems encountered in operations research would require substantial effort; second, it uses only the final solution from each previous problem, while our approach uses the full set of function evaluations; third, it requires that solutions from previous problems be evaluated under the current objective before they provide useful information, while our approach utilitizes this information directly without requiring additional expensive evaluations.

Finally, in the aerospace engineering literature, Gano, Renaud, and Sanders~\cite{grs04} studied the optimization of expensive-to-evaluate functions in a multi-fidelity setting and showed that warm starting optimization of an expensive high-fidelity objective using inexpensive evaluations of a low-fidelity model can reduce the number of high fidelity queries required.  This previous work is focused on engineering design problems in which computational codes of various expense and fidelities all model the same objective, while we focus on a sequence of optimization problems with different objectives, that are not ordered by fidelity, and do not necessarily have different computational costs.

\medskip

\noindent{\textbf{Outline of the Paper.}}
The problem is stated formally in Sect.~\ref{section_problem_def}. 
The warm start algorithm is described in Sect.~\ref{section_model_overall}.
The results of an experimental evaluation of the algorithm are presented in Sect.~\ref{section_experiments}.

\section{PROBLEM DEFINITION}
\label{section_problem_def}
We give a formal exposition of the setup. 
We indicate the current problem to be solved by $\Task_0$, and tasks that have been solved previously by $\Task_\ell$, for $\ell > 0$.  We also refer to these optimization problems as ``instances'' or ``tasks'' in the sequel.  Let $M$ be the number of previously solved problems, and define~$[M] := \{1,2,\ldots,M\}$ and~$[M]_0 := [M] \cup \{0\}$.

We suppose that evaluating~$\Task_\ell$, $\ell\ge0$, at input parameter~$x$ yields independently and normally distributed observations with mean~$f(\ell,x)$ and variance~$\lambda_\ell(x)$, i.e.\ $f(\ell,x)$ denotes the objective value of~$x$ for the current task.  We also suppose that accompanying this evaluation is an estimate of $\lambda_\ell(x)$.  This assumption of normally distributed noise accompanied by an estimate of the noise's variance is approximately valid when each evaluation is actually an average of many independent and identically distribution observations, whose sample variance provides an estimate of~$\lambda_\ell(x)$.   

We indicate the current optimization problem's feasible set by $\domain$.  This feasible set may be shared with previous problems, or previous problems may have different feasible sets.  We assume that $\domain$ is a compact subset of $\mathbb{R}^d$, and that membership in this set can be determined simply without evaluating an expensive function.  For example, $\domain$ might be a hyper-rectangle or a simplex. 

Our goal is to optimize the current task, i.e., to find ~$\argmax_{x \in \domain} f(0,x)$.  To support us in this task we have evaluations of previous tasks, in the form of a collection of tuples $(\ell,x,y_\ell(x),\lambda_\ell(x))$, where $y_\ell(x)$ is the sum of $f(\ell,x)$ and independent normally distributed noise with variance $\lambda_\ell(x)$.  Typically these would be obtained from runs of the optimization algorithm we propose, or some other optimization algorithm, when solving these previous tasks.  Evaluations obtained for other simulation studies can also be included.

During optimization for the current task, we assume that previous tasks cannot be evaluated.  We make this assumption because maintaining the ability to evaluate previous tasks may present greater practical engineering challenges than simply storing previous evaluations, if previous tasks used different versions of the simulation software, or required access to large historical datasets.  In contrast, under the assumptions made below, previous evaluations do not consume a great deal of memory, and can be stored simply in a flat file or database.

We make a few additional assumptions that should be met for our proposed methodology to be useful in comparison to other methods. 
\begin{enumerate}
\item First, we suppose that evaluations of $\Task_\ell$ do not provide first- or second-order derivative information.  If derivative information is available, it would be more productive to instead use a gradient-based method.
\item Second, we suppose that evaluations are expensive to obtain, limiting the number of evaluations that can be performed to at most a few hundred evaluations per problem.  When the number of evaluations possible grows substantially larger, the computational expense incurred by our method grows, decreasing its utility.
\item Third, we will model both the current problem's objective $x \mapsto f(0,x)$ and the collection of all objectives $(\ell,x) \mapsto f(\ell,x)$ with Gaussian processes, and so we assume that these objectives are amenable to Gaussian process modeling.  While Gaussian processes are flexible, and can handle a wide variety of objective functions, they require continuity of $x\mapsto f(\ell,x)$ for each $\ell$ to work well.
\item Fourth, while our mathematical approach does not make restrictions on $d$, our numerical experiments all assume that $d\le 8$.  Other work on Bayesian optimization suggests that performance degrades on high-dimensional problem, suggesting that the method we develop will provide value only for problems with $d\le 20$. 
\end{enumerate}

\section{WARM START ALGORITHM}
\label{section_model_overall}
The key idea of our approach is to use samples of the objectives from previous optimization problems to learn about the current optimization problem's objective.  To accomplish this, we use a Bayesian prior probability distribution that models the relationship between past objectives $x\mapsto f(\ell,x)$ and the current objective $f(0,x)$.
Thereby we begin our optimization of~$\Task_0$ with substantial information obtained from previous tasks, allowing us to find one of its optima faster.

More specifically, we place a Gaussian process prior on~$f$, which models the relationships between the tasks.  We emphasize that the Gaussian process prior is not just over $x \mapsto f(\ell,x)$ for a single $\ell$, as is typical in Bayesian optimization, but instead is over $(\ell,x) \mapsto f(\ell,x)$.  Modeling the objective jointly across tasks allows our approach to learn about the current task from previous ones.  This is described in Sect.~\ref{section_model}.

The Gaussian process prior utilizes a parametrized class of covariance functions: we describe in Sect.~\ref{section_hyper_estimation} how the parameters for the covariance function can be obtained in our problem setting.
These parameters describe the fundamental characteristics of the instances, and therefore are typically robust. We suggest to optimize them periodically rather than for every new task.

Then, using this Gaussian process prior over the collection of tasks, we apply Gaussian process regression to update our belief on~$f(0,\cdot)$ based on samples obtained from previously tasks.
Although this application of Gaussian process regression is relatively straightforward, we provide a detailed description for completeness in Sect.~\ref{section_gp_prior_and_post} showing how to calculate posterior probability distribution on $f(0,\cdot)$ given the prior from Sect.~\ref{section_model} and samples from previous tasks.

Finally, starting from this Gaussian process posterior over the current task's objective $f(0,\cdot)$, we use samples of $f(0,\cdot)$ to localize the optimum.  Points are chosen by our algorithm to sample by iteratively solving a sub-optimization problem, in which we find the point to evaluate that maximizes some acquisition function, sample this point, use it to update the posterior, and repeat until our budget of samples is exhausted.  We choose as our acquisition function the Knowledge Gradient (KG) proposed by Frazier, Powell, and Dayanik~\cite{FrPoDa_Correlated}; a succinct exposition is given in Sect.~\ref{section_algorithm}.

\paragraph*{The Algorithm \WSKG.} The proposed algorithm for warm starting Bayesian optimization proceeds as follows:
\begin{enumerate}
\item Using samples of previous tasks, and the prior distribution described in Sect.~\ref{section_model}:
	\begin{enumerate}
	\item Estimate hyper-parameters of the Gaussian process prior as described in Sect.~\ref{section_hyper_estimation}.
	\item Use the prior with the estimated hyper-parameters and the samples on previous tasks to calculate a posterior on $f(0,\cdot)$, as described in Sect.~\ref{section_gp_prior_and_post}.
	\end{enumerate}
\item Until the budget of samples on the current task is exhausted:
	\begin{enumerate}
	\item Choose the point $x$ at which to sample $f(0,x)$ next by maximizing the KG factor, as described in Sect.~\ref{section_algorithm}.
	\item Update the posterior distribution to include the new sample of $f(0,\cdot)$, as described in Sect.~\ref{section_gp_prior_and_post}, so that it includes all previous samples on the current and previous tasks.
	\end{enumerate}
\item Choose as our solution the point with the largest estimated value according to the current posterior.
\end{enumerate}

\subsection{Construction of the Prior}
\label{section_model}
We begin by describing the construction of the Gaussian Process prior,
which models the relationship between previous tasks and the current one.
Let~$\delta_\ell(x)$ denote the difference in objective value of any design~$x$ between~$\Task_0$ and~$\Task_\ell$: 
\begin{equation}
\delta_\ell(x) = f(\ell,x) - f(0,x).
\end{equation}
We construct our Bayesian prior probability distribution over $f$ by supposing that each~$\delta_\ell(\cdot)$ with $\ell>0$ is drawn from an independent Gaussian Process prior with mean function~$\mu_\ell$ and covariance kernel~$\Sigma_\ell$, where~$\mu_\ell(\cdot)$ and~$\Sigma_\ell(\cdot,\cdot)$ belong to some pre-specified parametrized class of functions. 
For example, we assume in our numerical experiments that $\mu_\ell(\cdot) = 0$, and that~$\Sigma_\ell(\cdot,\cdot)$ belongs to the Mat\'{e}rn covariance functions,  
a popular class of stationary kernels (see chapter~4.2 in Rasmussen and Williams~\cite{rw06} for details).
It would also be possible to set the prior mean $\mu_\ell(\cdot)$ to a constant.
Hyper-parameters determining the specific choices for $\mu_\ell$ and $\Sigma_\ell$ from their respective parameterized class can be fit from data in an automated process, as we describe in more detail in Sect.~\ref{section_hyper_estimation}.

This $\delta_\ell(x)$ models the difference between previous tasks and the current task.  By using knowledge of the typical values of $\delta_\ell(x)$, inferred from the hyper-parameters and from evaluations of $f(\ell,x')$ and $f(0,x'')$ for $x'$ and $x''$ nearby to $x$, we can learn about $f(0,x)$ from knowledge of $f(\ell,\cdot)$.

We also suppose that $f(0,x)$ is drawn from another independent Gaussian process with mean function $\mu_0$ and covariance kernel $\Sigma_0$, which also belong to some pre-specified parameterized class of functions.  As with $\mu_\ell$ and $\Sigma_\ell$ for $\ell>0$, these parameters may also be chosen automatically as described in Sect.~\ref{section_hyper_estimation}.

This specification of the (joint) distribution of $(\ell,x) \mapsto \delta_\ell(x)$ and $x \mapsto f(0,x)$, together with the specification of $f(\ell,x)$ in terms of these quantities, defines a joint Gaussian process probability distribution over $f$.  This is because each $f(\ell,x)$ is a linear function of quantities that are themselves jointly Gaussian.  By specifying the mean function and covariance kernel of this Gaussian process, which we call $\mu$ and $\Sigma$ respectively, we may leverage standard software for Gaussian process regression.

We calculate the functions $\mu$ and $\Sigma$ as follows.
For the ease of notation let~$\delta_0(x) = 0$ for all~$x \in \domain$, then we can write~$f$ as
$f(\ell,x) = f(0,x) + \delta_\ell(x)$.
Then, for $\ell \in [M]_0$, we calculate the mean function as,
\begin{equation*}
\mu(\ell,x)  = \E\left[f(\ell,x)\right]
 = \E\left[f(0,x) + \delta_\ell(x)\right]
 = \mu_0(x) + \E\left[\delta_\ell(x)\right]
 = \mu_0(x),
\end{equation*}
since~$\E\left[\delta_\ell(x)\right] = 0$.
For~$\ell,m \in [M]_0$ and~$x,x' \in \domain$, the covariance kernel is given by
\begin{align*}
\Sigma\left((\ell,x),(m,x')\right) & = Cov(f(\ell,x),f(m,x'))\\
& = Cov(f(0,x) + \delta_\ell(x), f(0,x') + \delta_m(x')),\\
& = Cov ( f(0,x), f(0,x')) + Cov(\delta_\ell(x), \delta_m(x'))\\
\intertext{and since the Gaussian processes~$\delta_\ell$ and~$\delta_m$ are independent iff~$\ell \neq m$ holds,}
& = \Sigma_0(x,x') + \I_{\ell,m} \cdot \Sigma_\ell(x,x'),
\end{align*}
where we use the indicator variable~$\I_{\ell,m}$ that is one if~$\ell = m$ and~$\ell \geq 1$, and zero otherwise.

\subsection{Estimation of the Hyper-Parameters}
\label{section_hyper_estimation}
When designing the method, we aimed for a wide applicability, and therefore do not assume that further information on the current optimization task~$\Task_0$ and its relationship with previous tasks is given explicitly. 
Rather we will estimate the important quantities from historical data in an automated fashion as follows.

Let~$\theta$ denote the parameters of the covariance function~$\Sigma_\ell$: for instance, in case of the Squared Exponential Kernel~$K(x,y) = \alpha_\ell \exp(\sum_{i = 1}^d \beta_{\ell,i} (x_i - y_i)^2)$ for~$x,y \in \mathbb{R}^d$ there are~$d+1$ parameters.
Since we believe that~$\theta$ changes little from instance to instance, we formulate a prior distribution based on previously optimal choices for~$\theta$: let~$p(\theta)$ be the probability under that prior. Moreover, our statistical model provides the likelihood of observations~$Y$ conditioned on the samples~$X$ and~$\theta$, denoted by~$p(Y \mid X, \theta)$. 
Thinking of~$\theta$ as a random variable, the probability of an outcome for~$\theta$ conditioned on the observations equals the product~$p(Y \mid X, \theta) \cdot p(\theta)$.
Thus, the method of maximum a posteriori (MAP) estimation the chooses the hyper-parameters that maximize this product, or equivalently as $\argmax_{\theta} \ln(p(Y \mid X, \theta)) + \ln(p(\theta))$, where~$\ln(\cdot)$ is the natural logarithm.
The optimal parameters can be computed efficiently via gradient methods. We refer to Sect.~5.4.1 in~\cite{rw06} for details.

\subsection{Acquisition Function}
\label{section_algorithm}
The proposed algorithm takes as input the historical data, i.e.\ samples obtained in the optimization on previous tasks, and an ``oracle access'' to the current task~$\Task_0$; that is, we consider the objective function a black-box. 
The actual optimization process is conducted in an iterative manner, where in each iteration a so-called acquisition function selects a point~$x$ to sample. 
A wide range of these functions has been proposed in the literature. We suggest using the Knowledge Gradient~\cite{FrPoDa_Correlated}: this greedy strategy is known to be one-step (Bayes) optimal and moreover converges to an optimal solution in settings where a finite number of alternatives is given with a multivariate normal prior.

From the practical point of view, the main advantage of the Knowledge Gradient is that has been shown to produce good results, but nonetheless can be computed very efficiently (see their paper for details on the implementation). For the sake of a self-contained presentation, we give a brief description of the method.

We assume that we have already observed~$n$ points for~$\Task_0$ and are now to decide the next sample.
Let~$\mu^{(n)}(x) = \E_n[f(0,x)]$, where~$\E_n$ denotes the expected value under the posterior distribution. 
Thus, based on our current knowledge, the best~$x \in \domain$ for task~$\Task_0$ has an expected objective value of~$\max_{x \in \domain}\mu^{(n)}(x)$. 
Note that this value can be computed with our information after observing the~$n$-th sample.
Then the Knowledge Gradient~(\KG) selects a design~$x^{(n+1)} \in \domain$ that maximizes
$\mathbb{E}_n\left[\max_{x' \in \domain} \mu^{(n+1)}(x') \;\middle|\; x^{(n+1)} = x\right] - \max_{x' \in \domain} \mu^{(n)}(x')$.
We obtain an approximation by replacing~$\domain$ by a discrete set of points, denoted by~$\discretedomain$:
\begin{equation}
\label{Eq_approx_voi_per_cost_def}
\mathbb{E}_n\left[\max_{x' \in \discretedomain} \mu^{(n+1)}(x') \;\middle|\; x^{(n+1)} = x\right] - \max_{x' \in \discretedomain} \mu^{(n)}(x').   
\end{equation}
Set~$\discretedomain$ can either be obtained by randomly sampling from~$\domain$, for example using a Latin Hypercube design to ensure an even coverage. If certain regions of~$\domain$ seem particularly relevant, then~$\discretedomain$ can represent it with higher resolution.
Note that the first summand of Eq.~(\ref{Eq_approx_voi_per_cost_def}), the conditional expectation for fixed~$x$, can be calculated from the posterior distribution:
recall that~$\mu^{(n)}$ is mean vector and let~$\Sigma^{(n)}$ denote the covariance matrix of the posterior distribution after observing~$n$ samples.
Define
$\tilde{\sigma}_{x'}(x) = \Sigma^{(n)}_{(0,x'),(0,x)} / (\Sigma^{(n)}_{(0,x),(0,x)} + \lambda_0(x))^{\frac{1}{2}}$.
It is well-known (e.g., see~\cite{kailath1968,FrPoDa_Correlated}) that the random variables~$\mu^{(n+1)}(x')$ and~$\mu^{(n)}(x') + \tilde{\sigma}_{x'}(x) \cdot Z$ have the same distributions conditioned on the previous samples. Here~$Z$ is a standard normal random variable. Thus we have that $\mu^{(n+1)}(x') \overset{d}{=} \mu^{(n)}(x') + \tilde{\sigma}_{x'}(x) \cdot Z$ 
holds for every~$x' \in \discretedomain$.

Based on that observation, we can either enumerate all~$x \in \discretedomain$ and pick a point that maximizes Eq.~(\ref{Eq_approx_voi_per_cost_def}), which may be useful if the cost of evaluating the task is very expensive (compared to the computational cost of the enumeration).
Or we utilize that the gradient of the~\KG can be computed efficiently (e.g., see~\cite{fxc11,sfp11,pwf16}) and perform a local search over~$\domain$, for instance using a multi-start gradient ascent method. In this case we would require that the noise functions~$\lambda_\ell(\cdot)$ are differentiable.

\section{EXPERIMENTAL EVALUATION}
\label{section_experiments}
We have evaluated the warm start algorithm on two benchmark suites.
The first set comprises variants of the two-dimensional Rosenbrock function that is a standard test problem in Bayesian optimization.

The second family of testbed instances in this evaluation are taken from the Assemble-To-Order (\ATO) problem that was introduced by Hong and Nelson~\cite{hong2006discrete}.
In this eight-dimensional simulation optimization scenario the task is to optimize the target of each item under a continuous-review base stock policy that manages the inventory of a production facility.  
In order to evaluate the objective value of a specific choice of the parameters, we run the simulator of Xie, Frazier, and Chick~\cite{simopt_ato} available on the SimOpt website.

The performance of the warm start algorithm (\WSKG) is compared to two traditional methods that do not benefit from previous optimizations of related tasks.

\medskip

\noindent\textbf{\EGO.}
The first baseline method is the well-known~\EGO algorithm of Jones, Schonlau, and Welch~\cite{JoScWe98}. \EGO is also a myopic algorithm  that iteratively selects one point to sample. \EGO's acquisition criterion is to optimize the expected improvement~(\EI): for this exposition suppose that the case that observations are noiseless and that~$y^\ast$ is the best objective value that was sampled so far. Let~$Y_x$ be the random variable that is distributed (normally) according to the posterior distribution of the objective value of some design~$x$, then we call
$\EI\left(x\right) := \E\left[\max\left\{Y_x - y^\ast, 0\right\}\right]$ 
the \emph{expected improvement} for~$x$. In each iteration \EGO greedily samples an~$x$ that maximizes this expectation.

\medskip

\noindent\textbf{\KG.} In order to assess the benefit of taking data on previous tasks into account, we also compare the new method to the ``vanilla version'' of the Knowledge Gradient~(\KG)~\cite{FrPoDa_Correlated}; see Sect.~\ref{section_algorithm}.

\medskip

\noindent\textbf{Experimental Setup.} 
In our evaluation all algorithms are given the same initial data points drawn randomly for each instance and each run. The warm start algorithm is additionally provided with one data set for the Rosenbrock instances and two for~\ATO: each set contains the samples collected during a single run on a related instance. A single run consists of~25 steps (also referred to as iterations) for each of the Rosenbrock instances and~50 steps for the Assemble-to-Order suite.

The hyper-parameters of the Gaussian Process affect how accurate the posterior distribution predicts the objective function of the current task. 
In accordance with our scenario, we optimized the hyper-parameters once for a single instance of each suite and then invoked~\WSKG with this fixed setting for all instances.
For the baseline methods~\EGO and~\KG we optimized these parameters for each instance in advance, thereby possibly giving them an edge over~\WSKG in this respect.

The plots below provide the mean of the gain over the initial solution for each iteration.
Error bars are shown at the mean plus and minus two standard errors, averaged over at least~100 replications.

\subsection{The Rosenbrock Family}
\label{section_rosenbrock}
The basis of these instances is the 2D Rosenbrock function 
\begin{align*} \label{eq:rosenbrock}
RB_1(x_1,x_2) & = (1-x_1)^2 + 100 \cdot (x_2 - x_1^2)^2,
\intertext{which subject to the following modifications:}
RB_2(x_1,x_2) &= RB_1(x_1,x_2) + .01 \cdot \sin(10 \cdot x_1 + 5 \cdot x_2)\\
RB_3(x_1,x_2) &= RB_1(x_1 + .01,x_2 - .005)\\
RB_4(x_1,x2) &= RB_2(x_1,x_2) + .01\cdot x_1
\end{align*}
Moreover, each function evaluation is subject to i.i.d.\ noise with mean zero and variance~$.25$.
The task is to optimize the respective function on the domain~$[-2,2]^2$.

\begin{figure}
\begin{minipage}[t]{0.48\textwidth}
\includegraphics[width=\linewidth]{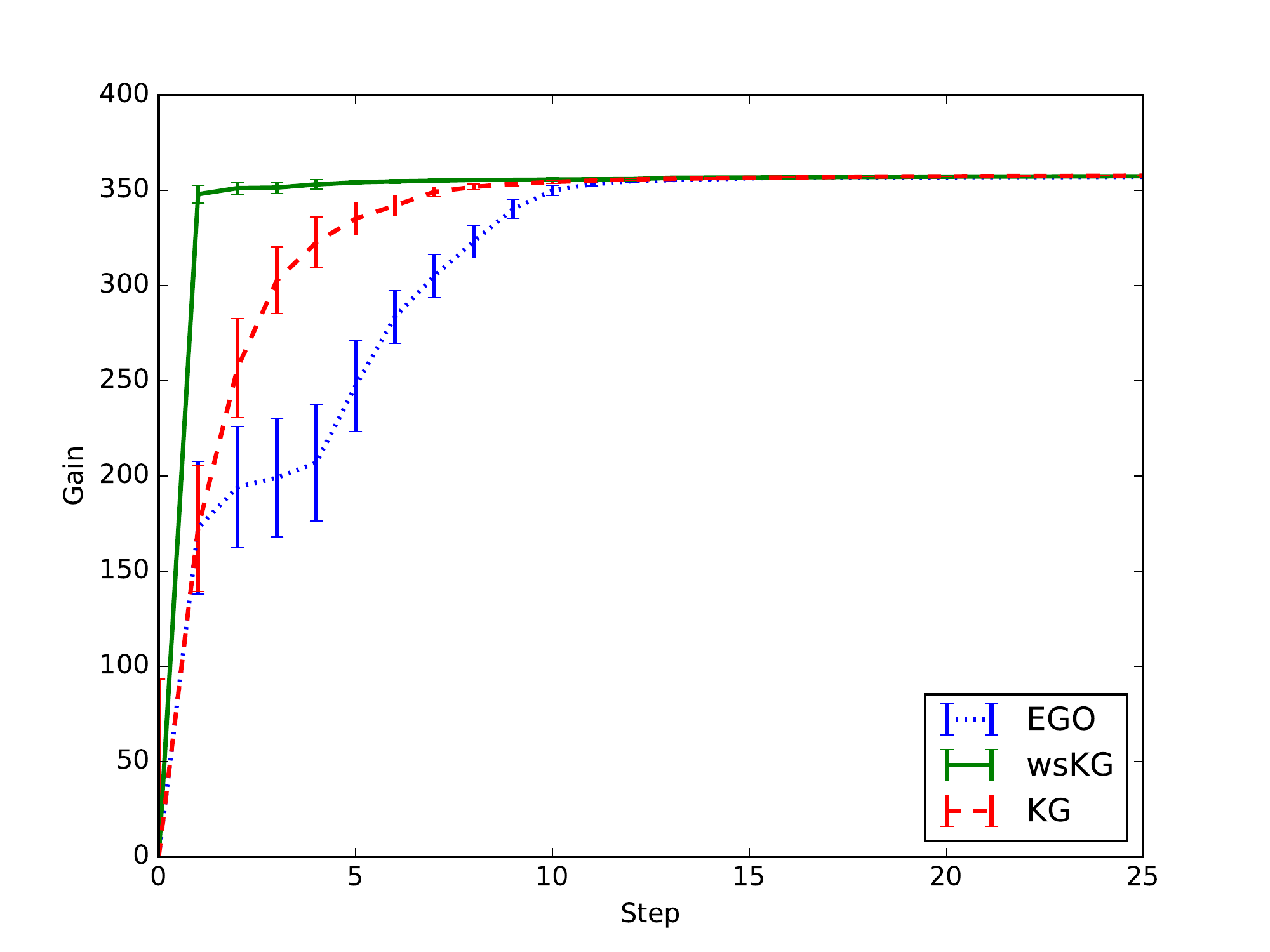}
\includegraphics[width=\linewidth]{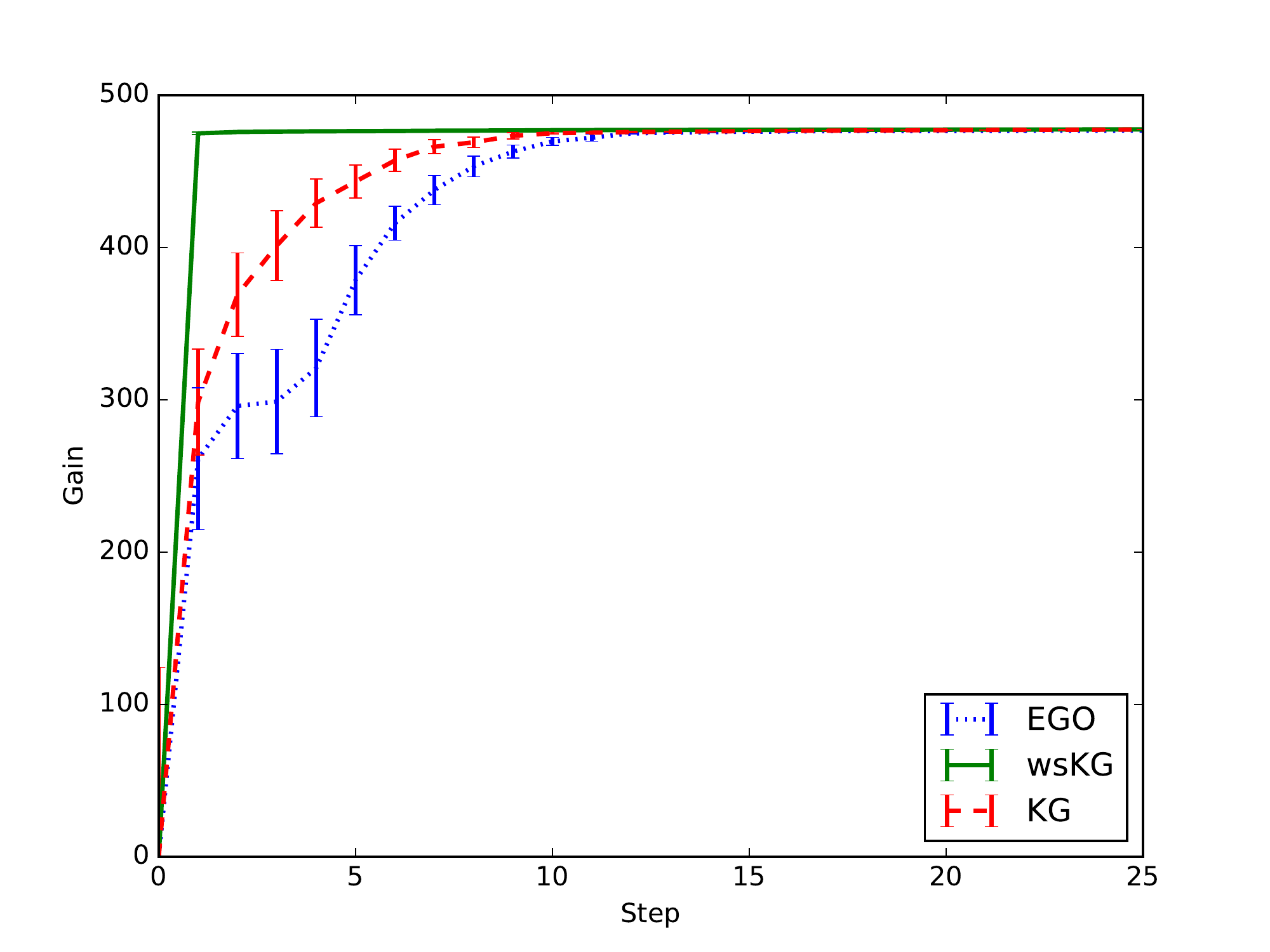}
\end{minipage}
\hfill
\begin{minipage}[t]{0.48\textwidth}
\includegraphics[width=\linewidth]{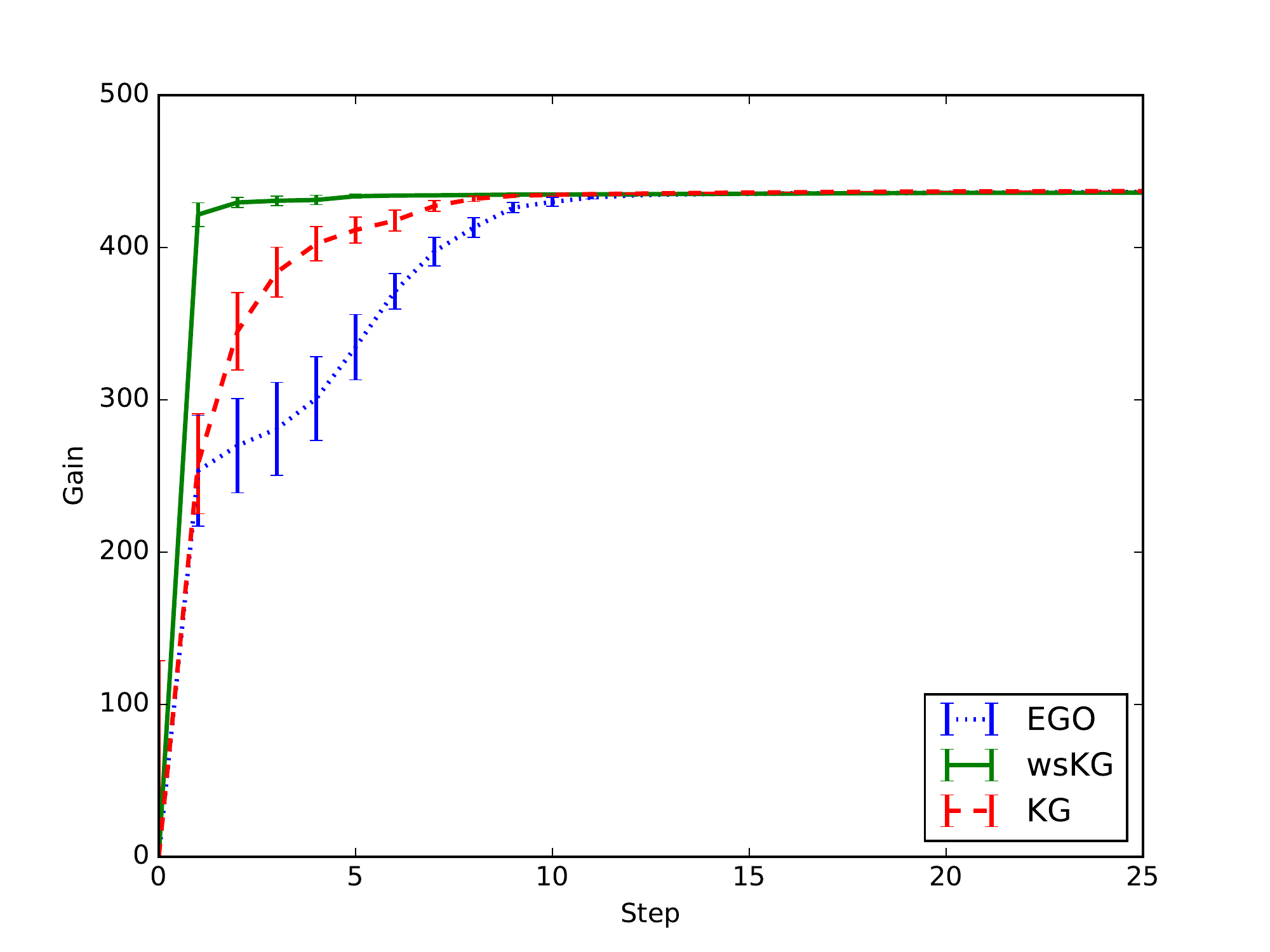}
\includegraphics[width=\linewidth]{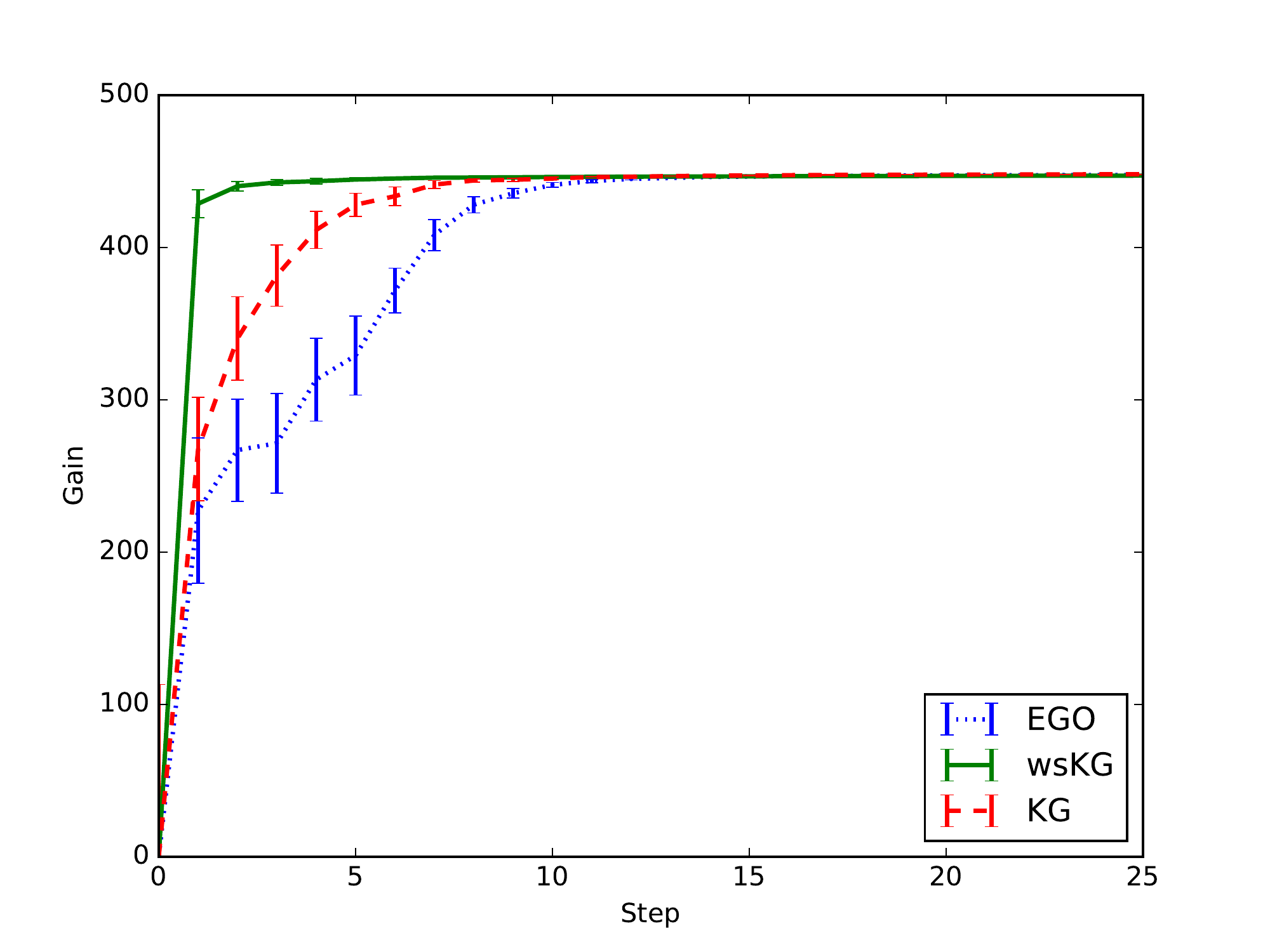}
\end{minipage}
\caption{(ul)~The basic Rosenbrock function~$RB_1$. (ur)~The Rosenbrock function~$RB_2$ with an additive sine. (bl)~The shifted Rosenbrock function~$RB_3$. (br)~The Rosenbrock function~$RB_4$ with an additive sine and a bias depending on~$x_1$.}
\label{fig_rb_1}
\end{figure}

The results are summarized in Fig.~\ref{fig_rb_1}: the warm start algorithm is able to exploit the knowledge from a single run on a related instance and samples better solutions than the other methods from the first iteration on. In particular, \WSKG obtains a near-optimal solution already after only one or two samples. \KG and~\EGO are distanced, with~\KG showing a superior performance to~\EGO.

\subsection{The Assemble-To-Order Benchmark}
In the Assemble-To-Order problem we consider a production facility that sells~$m$ different products that are assembled on demand whenever a customer request comes in.
Products are manufactured from a selection of~$n$ different items, where we distinguish for each product between \emph{key items} and \emph{non-key items}: when a product is ordered whose key items are all stocked, then it is assembled and sold. Otherwise the customer is lost. Non-key items are optional and used if available; they increase the revenue.
On the other hand, items kept in the inventory inflict \emph{holding cost}, therefore the facility would like to avoid overstocking.

The inventory is managed by a continuous-review base stock policy that sets a target stock base for each item and triggers a replenishment order if an item is requested.
The delivery time for supplies is normally distributed and varies over the items.
Moreover, the orders for products vary, too, and are modeled by~$m$ Poisson arrival processes.
The goal is to choose the target stock base used by the inventory management system so that the expected daily profit is maximized.

Hong and Nelson~\cite{hong2006discrete} proposed a setting (referred to as \ATO1) for~$n=8$ items,~$m=5$ products, and search domain~$[0,20]^8$.
We have created three additional instances based on \ATO1:

In the first time period~\ATO2 the forecast sees a change in the customer behavior: the expected demand for the two products that had been most popular before drops by~$3-5\%$. On the other hand, the popularity of two other products grows by up to~$5\%$. Additionally, the profit of some of the items increases by~$1-2\%$ on average. 

In the next period~\ATO3 the delivery of some of the items is delayed; the expected waiting time increases by about~$3\%$. Moreover, all products see a higher demand and the profits for several items increase.

In the final period~\ATO4 the production facility experiences holding costs rising by~$5\%$. Moreover, the profits of several items drop slightly.

\begin{figure}
\begin{minipage}[t]{0.48\textwidth}
\includegraphics[width=\linewidth]{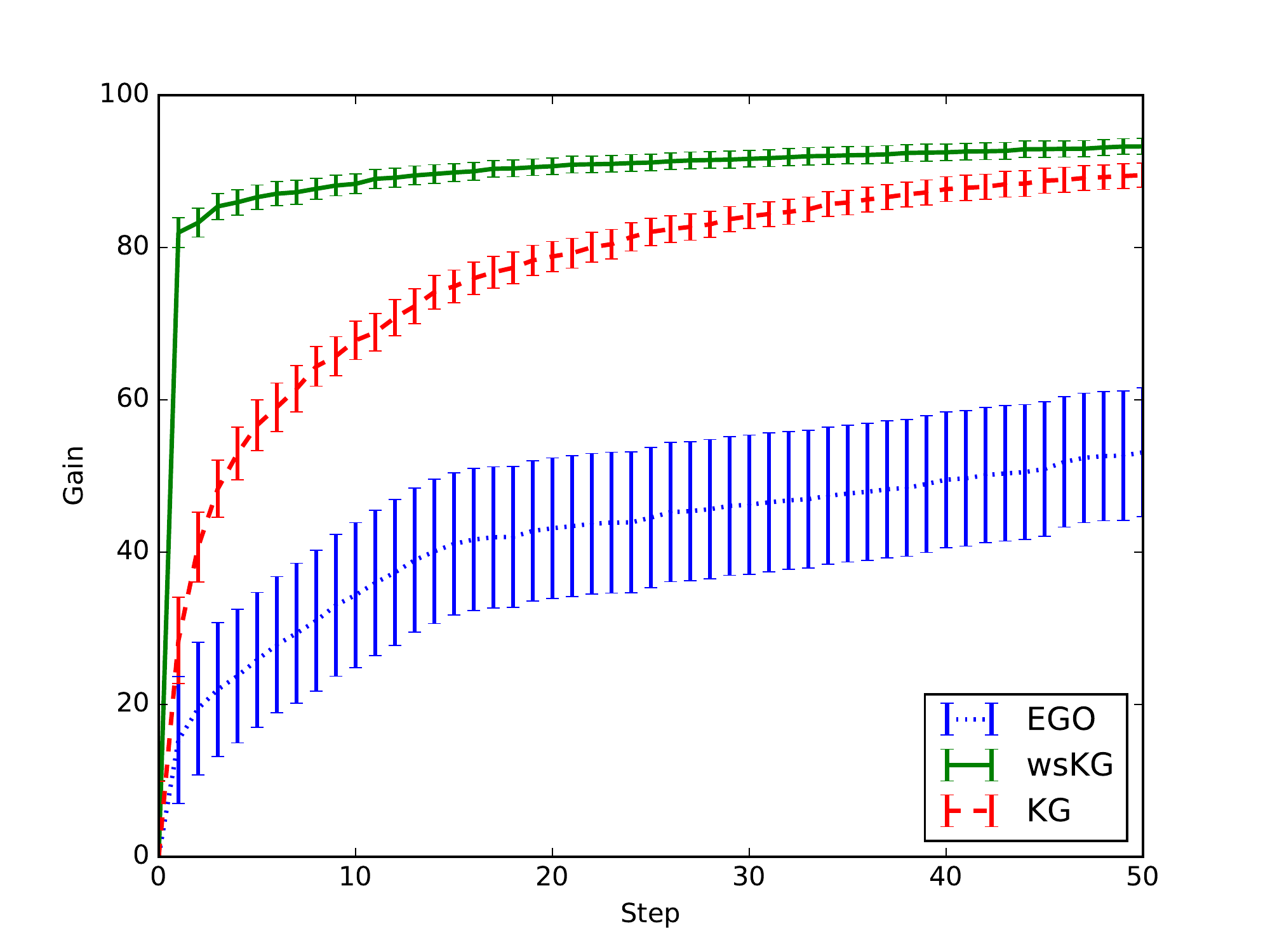}
\end{minipage}
\hfill
\begin{minipage}[t]{0.48\textwidth}
\includegraphics[width=\linewidth]{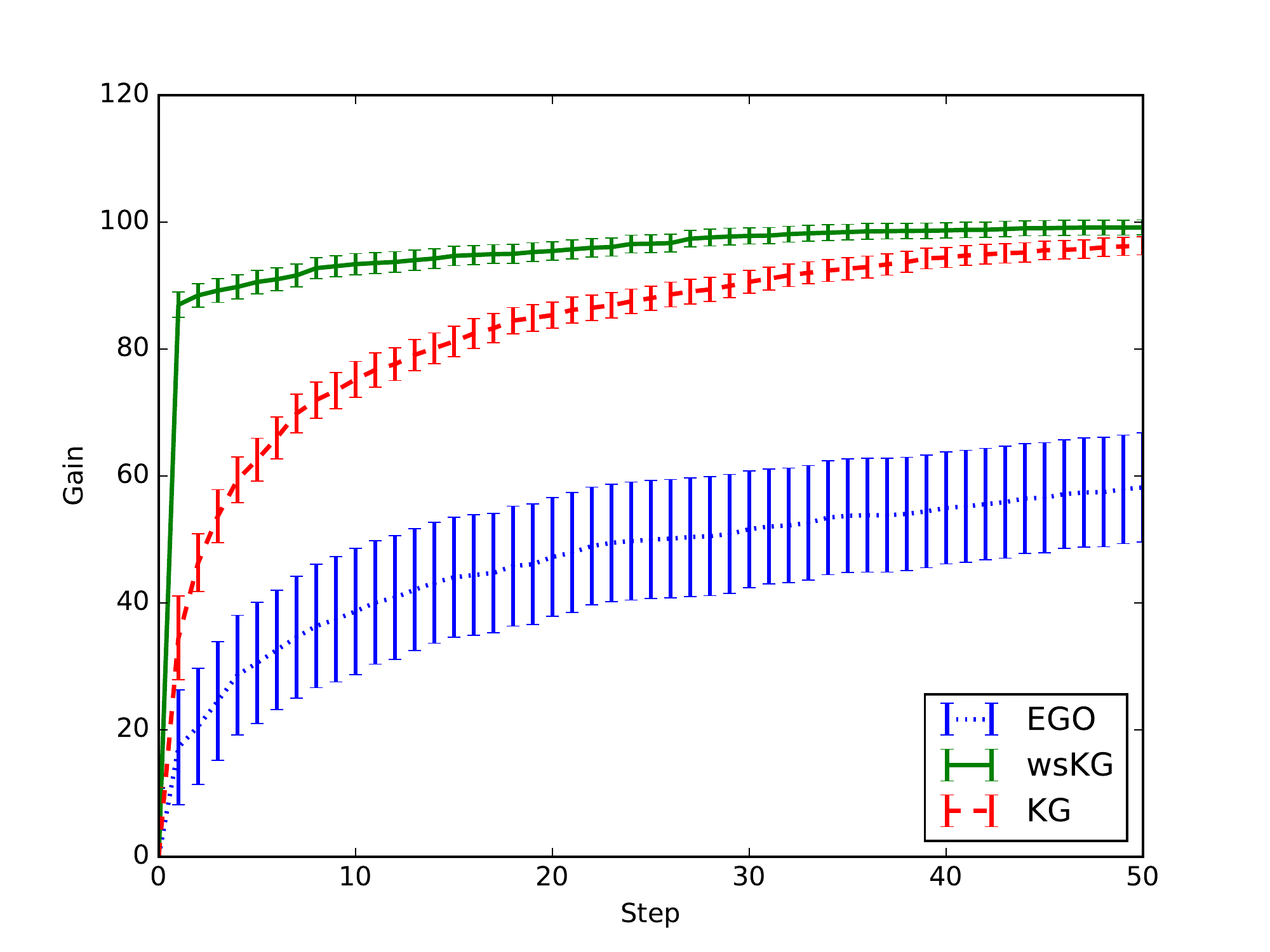}
\end{minipage}
\caption{(l) \ATO1 (r) \ATO2: All algorithms have the same initial data for the current problem. \WSKG has also access to samples of two runs on related instances, but its hyper-parameters are not optimized for the current instance.}
\label{fig_ato_1}
\label{fig_ato_2}
\end{figure}
\begin{figure}
\begin{minipage}[t]{0.48\textwidth}
\includegraphics[width=\linewidth]{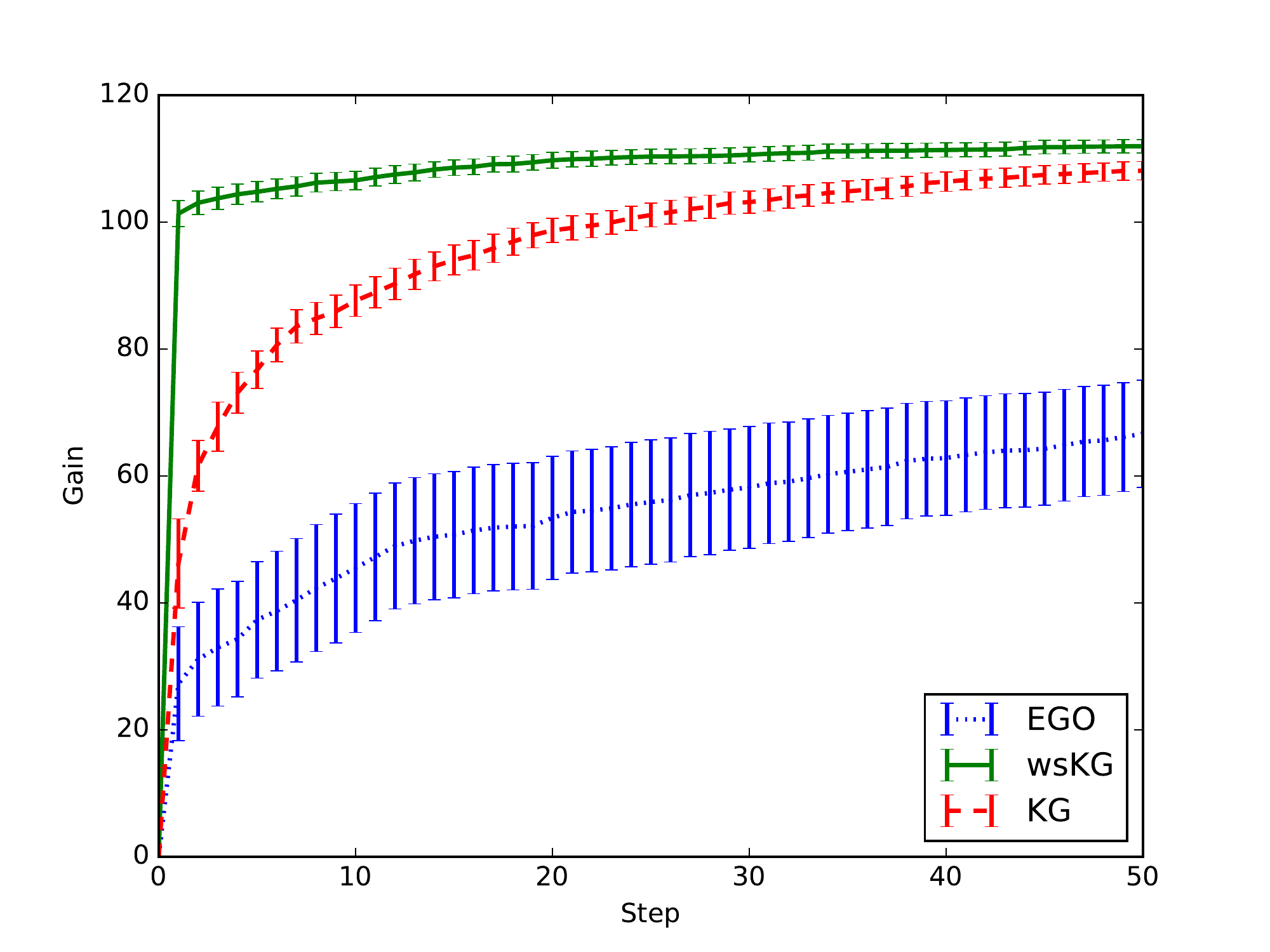}
\end{minipage}
\hfill
\begin{minipage}[t]{0.48\textwidth}
\includegraphics[width=\linewidth]{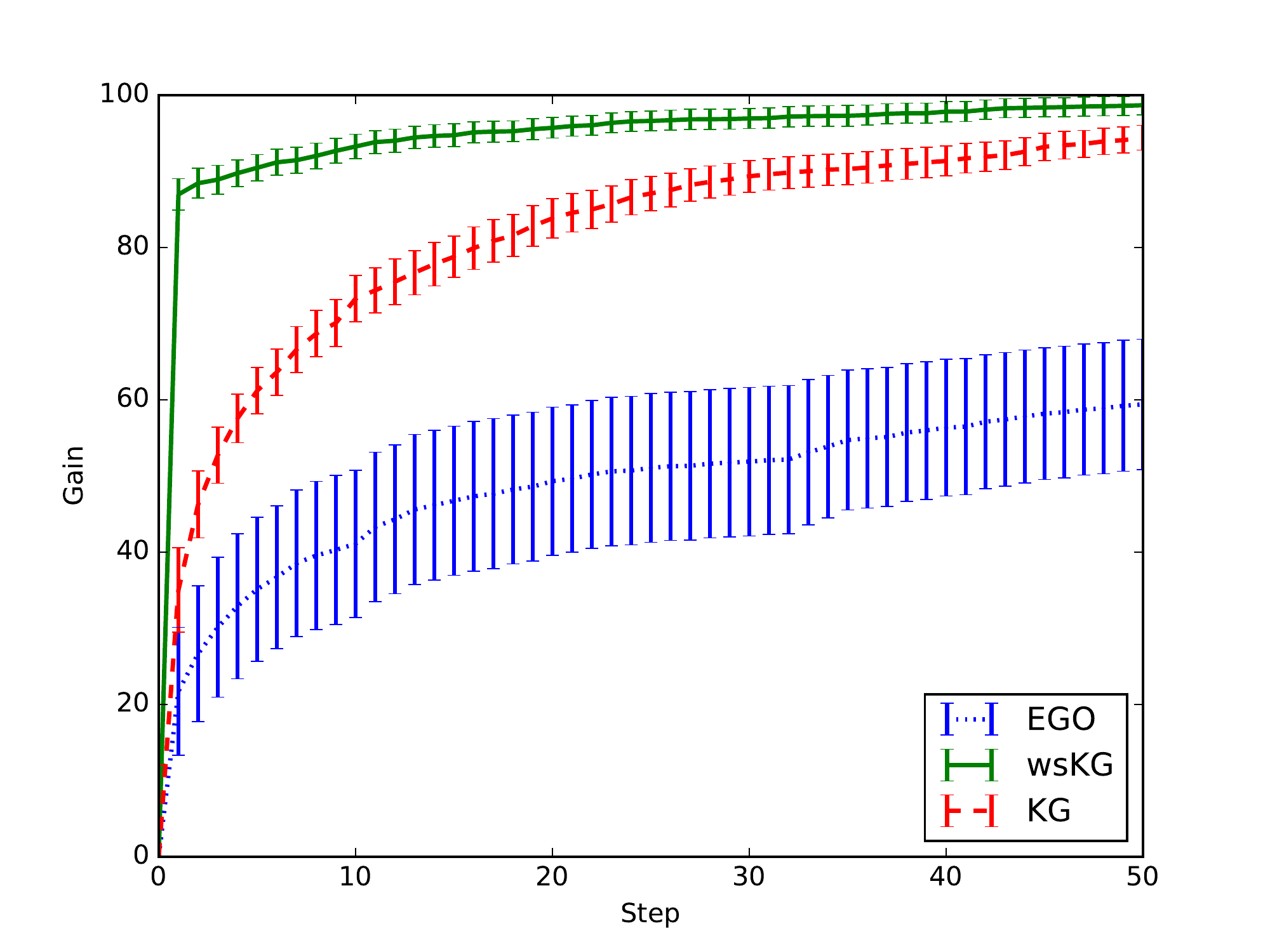}
\end{minipage}
\caption{(l) \ATO3 (r) \ATO4: \WSKG has received the samples of two runs on related instances.}
\label{fig_ato_3}
\label{fig_ato_4}
\end{figure}

When comparing the performance of the three algorithms given in Fig~\ref{fig_ato_1} and Fig~\ref{fig_ato_3} for the task of selecting the inventory levels used by the stocking policy, we see that \WSKG consistently performs significantly better than its competitors, achieving about~$95\%$ of the optimum after only ten samples.
Among the two distanced methods, the \KG policy achieves better solutions than~\EGO for the Assemble-To-Order problem. 
Looking closer, \KG's solution is typically about~$25\%$ away of the optimum after~10 steps, about~$15\%$ after~20 steps, and still about~$4\%$ after~50 steps, depending on the instance. \EGO achieves less than half of the optimum after~20 iterations, and only about~$50-60\%$ after~50 steps.
Notably, \EGO shows a large discrepancy between the performance of single runs; the error bar displays the mean averaged over~100 runs.

\section{CONCLUSION}
\label{section_conclusion}
We have proposed an algorithm to warm start Bayesian optimization and demonstrated its usefulness for an application in simulation optimization. The design of the algorithm is conceptually simple and computationally efficient.

A typical drawback of such Bayesian optimization methods is their ability to scale to larger input sizes.
For the chosen acquisition function, Poloczek, Wang, and Frazier~\cite{pwf16} recently suggested a novel parallelization that efficiently utilizes several multi-core CPUs. Thus, the acquisition function poses no longer the bottleneck of the approach if access to a computing cluster is provided.
However, in order to obtain the predictive distribution of the Gaussian Process model involves calculating the Cholesky decomposition to compute the inverse of a matrix, whose running time dependency on the number of previous observations is cubic. 

There are several approaches to address this and we believe it is worthwhile to combine one of them with the proposed method.
On the one hand, a possible direction is to replace the Gaussian Process by a scalable approximation. Several such techniques have been proposed lately, e.g., see~\cite{fwnns15,wn15} and the references therein.

A complementary approach to speed up the computations is to reduce the number of data points: instead of the whole data set, a ``sketch'' reduces the number of data points significantly, while the predictions made based on the sketch are nearly indistinguishable from the full dataset. Geppert et al.~\cite{gimqs15} proposed such an approximation and demonstrated its usefulness for applications in Bayesian linear regression. 
Another approach also based on a linear projection of the dataset to a subspace was given by Banerjee, Dunson, and Tokdar~\cite{bdt12}.

Another interesting direction is to extend our statistical model. We sketch two ideas. 
Recall that the statistical model in Sect.~\ref{section_model} describes the previous tasks as deviations from the current objective~$g(x)$, where each deviation is given by a Gaussian processes~$\delta_\ell$. In particular,~$\delta_\ell$ and~$\delta_{\ell'}$ are supposed to be \emph{independent} for~$\ell \neq \ell'$.
Alternatively, we could apply the formulation described in Sect.~2.2 of~\cite{pwf16} to additionally incorporate correlations among previous tasks.

A second approach is to suppose that all tasks are derived from an ``ideal'' problem. Then~$f(1,x) = g(x) = f(0,x) + \delta_1(x)$ gives the current task that we want to optimize, and~$f(\ell,x) = f(0,x) + \delta_\ell$ for~$\ell \geq 2$ are the previous tasks. Note that~$f(0,x)$ is not observable in this case and obtained via Gaussian process regression.
Both approaches can be implemented with little additional effort. Therefore, we have a collection of flexible models to choose from, depending on the characteristics of the optimization problem at hand.

\appendix

\section{PRIOR AND POSTERIOR DISTRIBUTION OF THE GAUSSIAN PROCESS}
\label{section_gp_prior_and_post}
For the sake of a self-contained exposition, we give the prior and the posterior of the Gaussian process. Both follow canonically (see~\cite{rw06} for an in-depth treatment).

Let~$X = \left((\ell^{(1)},x^{(1)}),(\ell^{(2)},x^{(2)}),\ldots,(\ell^{(n)},x^{(n)})\right)^T$ be a column vector of sample locations (for generality, assume that each pair belongs to~$[M]_0 \times \domain$).
Then the \emph{prior distribution} of~$f$ at~$X$ is given by
\begin{equation}
\label{Eq_prior}
\left(f(\ell^{(1)},x^{(1)}),\ldots,f(\ell^{(n)},x^{(n)})\right)^T \sim {\cal N}\left(\mu_0(X),K(X,X)\right),
\end{equation}
i.e.\ the prior distribution is multivariate normal.
Here we utilized that~$\mu\left(\ell,x\right) = \mu_0(x)$ for any~$\ell \in [M]_0$ and~$x \in \domain$ as shown above; $\mu_0(X)$ is a shorthand for the~$n$-dimensional column vector with~$\mu_0(X)_i = \mu_0(x^{(i)})$, and~$K\left(X,X\right)$ is the~$n \times n$ matrix with
$K\left(X,X\right)_{i,j} = \Sigma\left(\left(\ell^{(i)},x^{(i)}\right),\left(\ell^{(j)},x^{(j)}\right)\right)$
for~$i,j \in [n]$.

Next we describe the Gaussian Process regression to predict the latent function~$f(0,\cdot)$ that serves as our estimator for~$\Task_0(\cdot)$.
Upon sampling the points~$X$, we have observed~$Y := \left(y^{(1)},y^{(2)},\ldots,y^{(n)}\right)^T \in \mathbb{R}^n$ with 
\begin{equation*}
y^{(i)} = y_{\ell^{(i)}}\left(x^{(i)}\right) = f\left(\ell^{(i)},x^{(i)}\right) + \varepsilon^{(i)},
\end{equation*}
where~$\varepsilon^{(i)} \sim {\cal N}\left(0,\lambda_{\ell^{(i)}}\left(x^{(i)}\right)\right)$ for~$i \in [n]$.
Recall that~$\lambda_{\ell^{(i)}}\left(x^{(i)}\right)$ is the noise function that gives the variance when evaluating task~$\Task_{\ell^{(i)}}$ for design~$x^{(i)}$, and observe that~$\varepsilon^{(i)}$ and~$\varepsilon^{(i')}$ are independent (for~$i \neq i'$) conditioned on~$x^{(i)}$, $x^{(i')}$, and the respective noise functions.

Then the \emph{posterior} (or \emph{predictive}) \emph{distribution} of the latent function value~$f\left(0,x\right)$ is still multivariate normal (e.g., see Eq.~(A.6) on pp.~200 in~\cite{rw06}) with
\begin{align}
& f\left(0, x\right) \mid x, X, Y \sim\; {\cal N}\left(\mu^{n}(x),\sigma^{2,n}(x)\right) \quad\text{where} \nonumber\\
& \mu^{n}(x) = \mu_0\left(x\right) + K\left(x,X\right) \left[K\left(X,X\right) + D(X)\right]^{-1} \left( Y - \mu_0\left( X \right) \right) \label{Eq_posterior}\\
& \sigma^{2,n}(x) = \Sigma_0(x,x) - K\left(x,X\right) \left[K\left(X,X\right) + D(X)\right]^{-1} K\left(x,X\right)^T, \nonumber
\end{align}
where~$K\left(x,X\right)$ is defined as the row vector of length~$n$ with
$K\left(x,X\right)_i = \Sigma\left(\left(0,x\right),\left(\ell^{(i)},x^{(i)}\right)\right)$
for~$i \in [n]$, and~$D(X)$ is the~$n \times n$ diagonal matrix with
$D(X)_{i,i} = \lambda_{\ell^{(i)}}\left(x^{(i)}\right)$.
As usual,~$Q^{-1}$ denotes the inverse of some~$n \times n$ matrix~$Q$ under matrix multiplication, i.e.\ $Q Q^{-1} = I_n$ holds where~$I_n$ is the~$n \times n$ identity matrix.

\bibliographystyle{alpha}			\bibliography{multifidelity}		
\end{document}